\documentclass[journal]{IEEEtran}
\usepackage{graphics} % for pdf, bitmapped graphics files
\usepackage{epsfig} % for postscript graphics files
\usepackage{mathptmx} % assumes new font selection scheme installed
\usepackage{times} % assumes new font selection scheme installed
\usepackage{amsmath} % assumes amsmath package installed
\usepackage{amssymb}  % assumes amsmath package installed
\usepackage{subfigure}
\usepackage{gensymb}

\ifCLASSINFOpdf
\else
\fi
\begin{document}
\title{Integrating Inter-vehicular Communication, Vehicle Localization, and a Digital Map for Cooperative Adaptive Cruise Control with Target Detection Loss}

\author{Yuan~Lin,~\IEEEmembership{Member,~IEEE,}
        and~Azim~Eskandarian,~\IEEEmembership{Senior~Member,~IEEE}% <-this % stops a space
% \thanks{ is  Autonomous Systems and Intelligent Machines Lab in the Mechanical Engineering Department, Virginia Tech, Blacksburg, VA, 24061 USA e-mail: yuanlin@vt.edu.}% <-this % stops a space
\thanks{Dr. Azim Eskandarian is a full professor, the Director of the Autonomous Systems and Intelligent Machines (ASIM) Lab, and the Department Head of Mechanical Engineering at Virginia Tech, Blacksburg, VA 24061, USA (e-mail: eskandarian@vt.edu).}
\thanks{Dr. Yuan Lin was a postdoctoral research associate of the ASIM Lab in Mechanical Engineering Department at Virginia Tech when the research work was conducted (e-mail: yuanlin@vt.edu).}% <-this % stops a space
% \thanks{Manuscript received May 18, 2018; revised August 26, 2018.}
}


\markboth{IEEE Transactions on Intelligent Transportation Systems}
%\markboth{IEEE Transactions on Intelligent Transportation Systems, ~Vol.~14, No.~8, August~2015}%
{Shell \MakeLowercase{\textit{et al.}}: Bare Demo of IEEEtran.cls for IEEE Journals}



\maketitle

\begin{abstract}

% not use range sensor because of stereo cameras

Adaptive Cruise Control (ACC) is an Advanced Driver Assistance System (ADAS) that enables vehicle following with desired inter-vehicular distances. Cooperative Adaptive Cruise Control (CACC) is upgraded ACC that utilizes additional inter-vehicular wireless communication to share vehicle states such as acceleration to enable shorter gap following. Both ACC and CACC rely on range sensors such as radar to obtain the actual inter-vehicular distance for gap-keeping control. The range sensor may lose detection of the target, the preceding vehicle, on curvy roads or steep hills due to limited angle of view. Unfavourable weather conditions, target selection failure, or hardware issue may also result in target detection loss. During target detection loss, the vehicle following system usually falls back to Cruise Control (CC) wherein the follower vehicle maintains a constant speed. In this work, we propose an alternative way to obtain the inter-vehicular distance during target detection loss to continue vehicle following. The proposed algorithm integrates inter-vehicular communication, accurate vehicle localization, and a digital map with lane center information to approximate the inter-vehicular distance. In-lab robot following experiments demonstrated that the proposed algorithm provided desirable inter-vehicular distance approximation. Although the algorithm is intended for vehicle following application, it can also be used for other scenarios that demand vehicles' relative distance approximation. The work also showcases our in-lab development effort of robotic emulation of traffic for connected and automated vehicles.
\end{abstract}


% \fi
\IEEEpeerreviewmaketitle

\section{Introduction}

ACC systems are commercially available vehicle following systems that allow a vehicle to adjust its speed to maintain a desired distance or time gap between the vehicle itself and its preceding vehicle \cite{eskandarian2012}. The time gap setting is a popular choice as it allows the inter-vehicular distance to increase linearly with the follower vehicle's speed, which abides with safety concerns. An ACC system usually includes two components: one is obtaining the inter-vehicular distance via sensors and their algorithms, and the other is gap-keeping feedback control \cite{liang1999optimal,naranjo2003adaptive,luo2010model,ganji2014adaptive}. As inter-vehicular connectivity is introduced \cite{blum2004}, CACC systems are developed such that vehicles follow one anther in a cooperative manner \cite{dey2016,zheng2016stability}. An exemplary CACC system builds upon ACC feedback control and adds additional feedforward control which utilizes the preceding vehicle's acceleration received wirelessly as the feedforward input \cite{naus2010}. It has been demonstrated that CACC systems maintain shorter gap following compared to ACC \cite{milanes2014cooperative,lin2017}.

Many automotive ACC systems utilize sensors such as radar, lidar, or stereo cameras to detect the preceding vehicle and obtain the inter-vehicular distance for gap-keeping control \cite{widmann2000comparison}. These sensors usually have a limited angle of view (except for 360$\degree$ view sensors) and may lose the target detection on curvy roads or steep hills \cite{ahmed2005object,sudou2006adaptive,engelman2001adaptive}. The target detection loss may also occur when the target selection algorithm of the sensor fails to differentiate the preceding vehicle from nearby vehicles in adjacent lanes. With unfavorable weather conditions such as fog, the sensors may also lose the target due to low reflectance \cite{austin1987relation,difranco2004radar}. In addition, hardware problems might happen which could lead to sensor failure. During these target detection loss scenarios, ACC usually falls back to CC such that the follower vehicle keeps a constant speed until the sensor detects the preceding vehicle again \cite{winner2015handbook}. The ACC target detection loss is evident in the system development and experienced by users. Due to such limitations, ACC is an assistance instead of a safety system and requires drivers' full attention at all time.

Vehicle-to-everything communication (V2X) which enables connectivity among vehicles, infrastructure, and pedestrians is a major trend of transportation revolution that will improve transportation mobility and safety. With inter-vehicular communication, vehicles can share acceleration for the CACC and share positions for blind spot warning. With vehicle-to-infrastructure communication, vehicles can perform communication-based highway merging \cite{rios2017automated}, eco-routing \cite{elbery2015eco}, and cooperative intersection control \cite{lee2012development}. Different countries may have different communication standards for V2X such as the Dedicated Short Range Communication (DSRC) in the US \cite{kenney2011dedicated} and Cellular-V2X in China \cite{chen2017vehicle}. 5G mobile network technology which is currently under development and deployment will provide much faster and more reliable wireless communication for V2X \cite{andrews2014will}.

Automated driving is another trend of transportation revolution that reduces human drivers' driving tasks. The Society of Automotive Engineers has characterized five levels of driving automation which ranges from ``No Automation'' to ``Full Automation''. ACC systems fall into the second level which is ``Driver Assistance''. The higher the automation level, the more sophisticated the technology. One of the technological areas for highly automated driving is vehicle localization \cite{bresson2017simultaneous}. A lot of automated driving tasks such as path following control and collision avoidance would require vehicle localization. The accuracy requirement for vehicle localization is at the centimeter scale due to performance and collision avoidance concerns \cite{vivacqua2018self}. Accurate vehicle localization is achieved through information fusion by fusing data from different sources which may include Global Positioning System (GPS), Inertial Measurement Unit (IMU), odometry, camera, Lidar, and a high-definition (HD) digital map \cite{levinson2007map,levinson2010robust,gu2016gnss}. Recent studies also investigate collaborative localization which adds communicated information in the data fusion to provide desirable localization results \cite{shen2017optimization}.

The purpose of this work is to address the problem of vehicle following when the range sensor loses detection of the target preceding vehicle. We propose a method the can approximate the inter-vehicular distance using the essential functions of connected and automated vehicles which include inter-vehicular communication and vehicle localization. The proposed algorithm is an alternative way of obtaining inter-vehicular distances as opposed to directly using range sensors. As a proof of concept, we implement and validate the proposed algorithm on autonomous mobile robots for CACC robot following. In-lab experiments have demonstrated that the proposed algorithm provides desirable inter-robot distance approximation. We conclude that the proposed algorithm is a viable solution for vehicle following during target detection loss in real-world driving.

The rest of this paper is organized as follows: Section II introduces the CACC problem formulation and control system design. Section III details the proposed algorithm for inter-vehicular distance approximation. Section IV documents the setup of our robotic emulation of traffic for CACC robot following experiments. Section V shows the CACC robot following experiment results with the proposed inter-vehicular distance approximation algorithm. Section VI draws the conclusion of this work.


%\IEEEPARstart{T}{his} demo file is intended to serve as a ``starter file''
%for IEEE journal papers produced under \LaTeX\ using
%IEEEtran.cls version 1.8b and later.
%% You must have at least 2 lines in the paragraph with the drop letter
%% (should never be an issue)
%I wish you the best of success.

\section{CACC Problem Formulation}

The problem that we are addressing is vehicle following with target detection loss. The vehicle following system can be either CACC or ACC. In this work, CACC is used since CACC has been proved in the literature to successfully enable shorter gap following as compared to ACC \cite{milanes2014cooperative,lin2017}. In this section, we focus on the mathematical formulation and an exemplary control system design of CACC while also introducing the target detection loss issue.

\begin{figure}[htbp]
\centering
\includegraphics[width=3.4in]{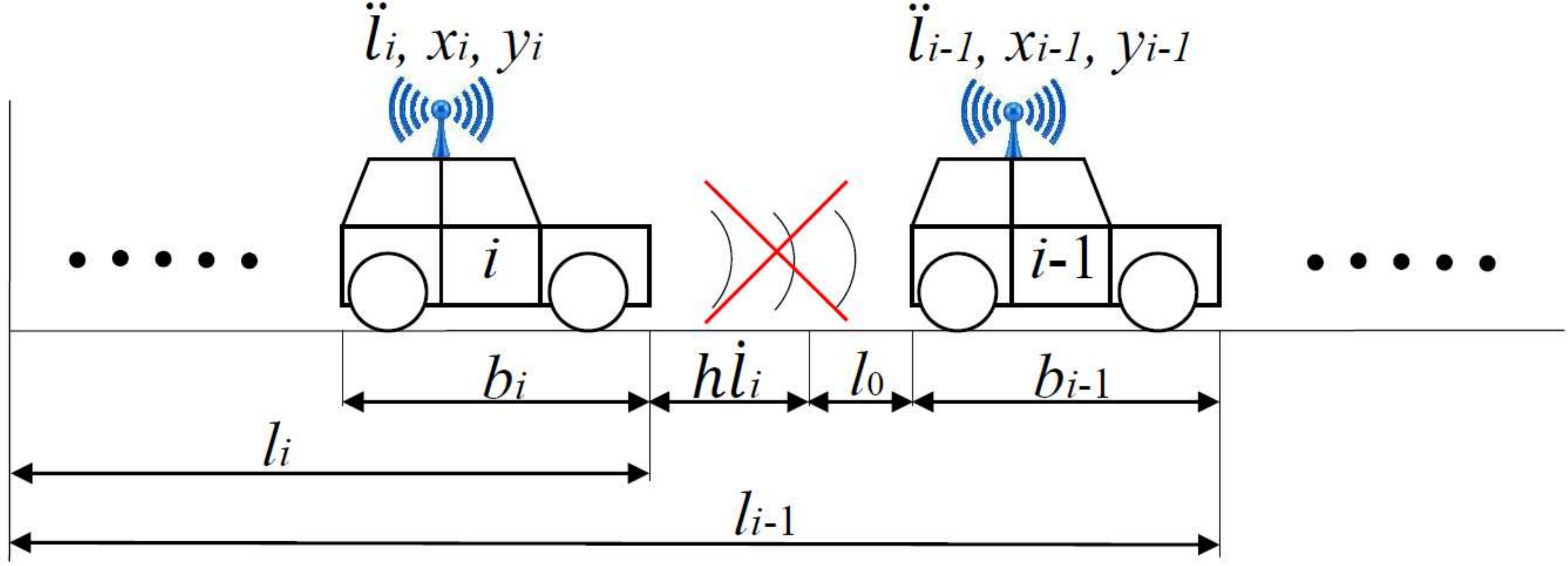}
\caption{Schematic for CACC vehicle following with target detection loss.}
\label{fig:schematic}
\end{figure}

Fig.~\ref{fig:schematic} shows a schematic of CACC vehicle following with target detection loss. The longitudinal distances traveled by the preceding $i-1$ and following $i$ vehicles (measured from the vehicle front bumper) are denoted as $l_{i-1}$ and $l_i$, respectively. Mathematically, the actual distance between these two vehicles is $l_{i-1} - l_i - b_{i-1}$ where $b_{i-1}$ is the body length of the preceding vehicle $i-1$. This actual inter-vehicular distance is obtained by range sensors such as radar in actual implementation. The desired distance is dictated by a constant time gap setting and is computed as $h \dot{l}_i + l_0$ where $h$ is the constant time gap, $\dot{l}_i$ is the follower vehicle's velocity, and $l_0$ is a standstill safety distance. The objective of CACC vehicle following control is to minimize the gap-keeping error between the actual and desired distances, i.e., $e_i = l_{i-1} - l_i - b_{i-1} - (h \dot{l}_i + l_0)$.

In Fig.~\ref{fig:schematic}, the symbols $x_{i-1}$ and $y_{i-1}$ are the horizontal and vertical positions of the preceding vehicle, respectively; the symbols $x_i$ and $y_i$ are the horizontal and vertical positions of the follower vehicle, respectively. These positions can be obtained through vehicle localization. These positions are cartesian coordinates that are used for the inter-vehicular distance approximation which is described in the next section.

\begin{figure}[htbp]
\centering
\includegraphics[width=3in]{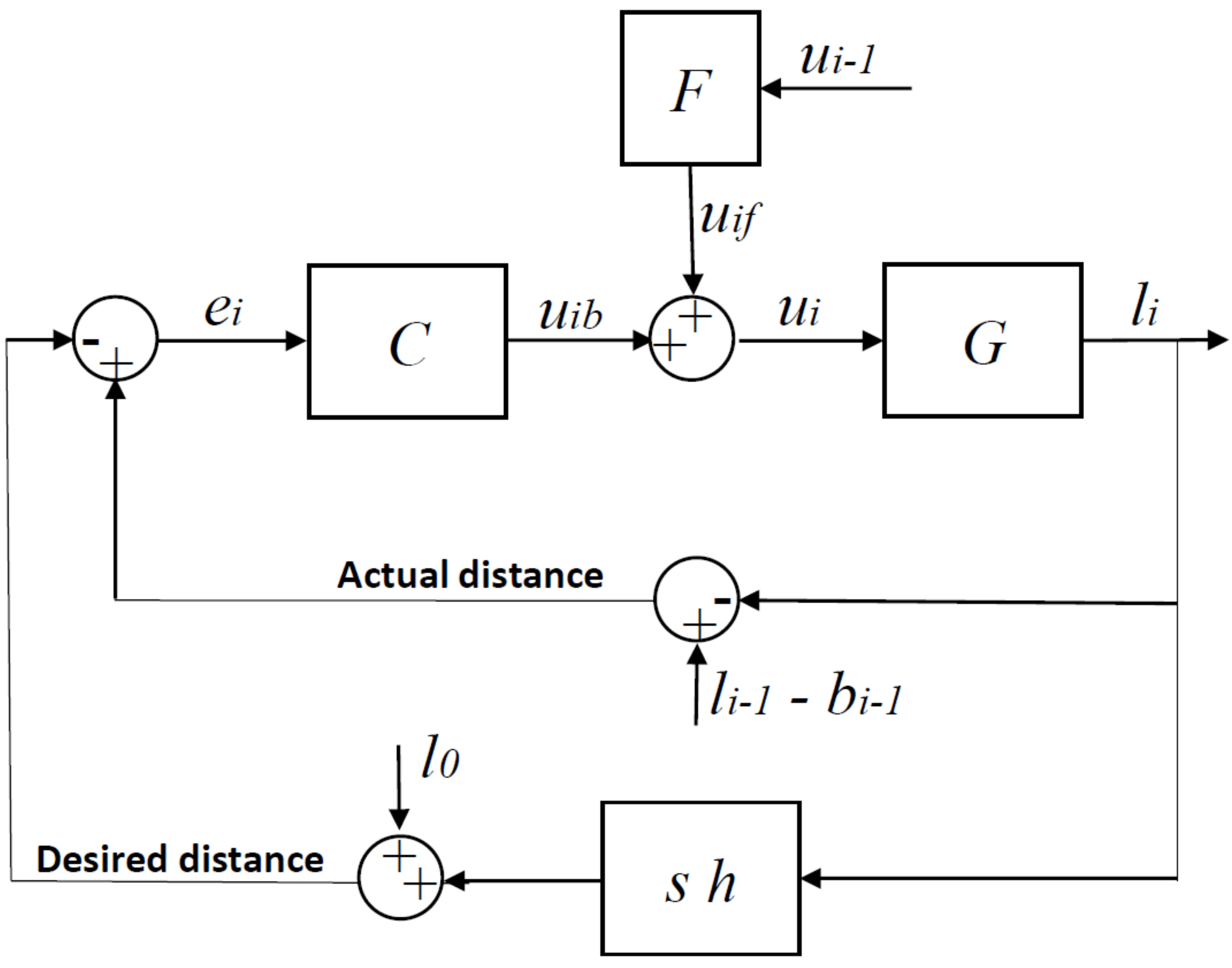}
\caption{Block diagram of the CACC system.}
\label{fig:cacc}
\end{figure}

Fig.~\ref{fig:cacc} shows the block diagram of an exemplary CACC system \cite{naus2010}. This CACC system consists of two portions: one is the traditional ACC feedback control that minimizes the error between the actual and desired distances; the other is the feedforward control that utilizes the acceleration of the preceding vehicle received through wireless communication. The feedforward control portion has a feedforward filter $F$ that filters the acceleration of the preceding vehicle. To derive the analytical expression of the feedforward filter $F$, the Laplace Transform of the error is obtained and set to zero $E_i = 0$ to obtain $F = 1/((1+sh)s^2 G)$. For the detailed derivation, readers can refer to \cite{naus2010} or our previous work \cite{lin2017experimental}. The feedforward input is supposed to be the instantaneous actual acceleration of the preceding vehicle. However, the actual acceleration that can be obtained through either IMU or wheel encoders can be noisy. Thus, we use the target acceleration of the preceding vehicle $u_{i-1}$ as the feedforward input.

The feedback control portion of the CACC system has a feedback controller $C$ that minimizes the gap-keeping error. The feedback controller that we use is a Proportional-Derivative (PD) controller \cite{dey2016}. It has been demonstrated in \cite{naus2010} that such CACC system design can realize vehicle following with string stability at a small time gap $h$ = 0.6 seconds.

% instead of a Proportional-Integral-Derivative (PID) controller. Generally speaking, the Integral part (I) of a PID controller is needed to eliminate the steady-state error of a control system. However, for either ACC or CACC vehicle following systems, there is actually no steady-state error with a PD controller. As this is non-trivial, we provide the analytical proof below using the Final Value Theorem.

%(Discuss the communication topology!!  Each vehicle broadcasts its position values to its neighboring vehicles through wireless communication. But only the follower vehicles use them.)

\section{Inter-vehicular Distance Approximation}

\begin{figure}[htbp]
\centering
\includegraphics[width=3.4in]{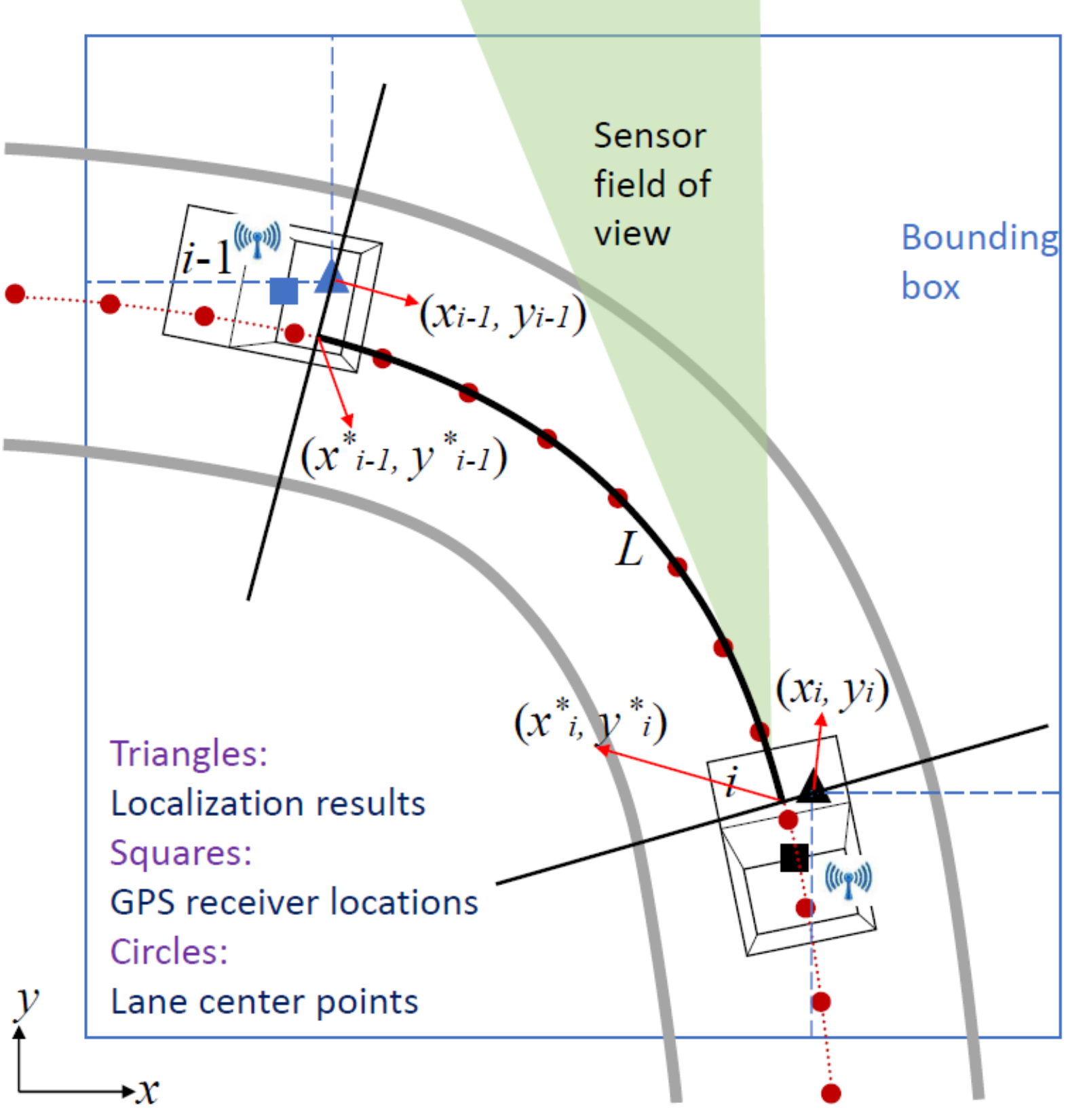}
\caption{Schematic for the inter-vehicular distance approximation algorithm.}
\label{fig:distance}
\end{figure}

This section illustrates the inter-vehicular distance approximation algorithm which is the core of this work. The algorithm requires accurate vehicle localization to obtain vehicle positions, inter-vehicular wireless communication to share those positions, and a digital map with lane center points. Fig.~\ref{fig:distance} shows the method of the inter-vehicular distance approximation. The main idea of this method is to obtain the inter-vehicular distance based on vehicles' projected positions on a quadratic curve that fits the lane center points.

% (not GPS receiver location?)

Firstly, each vehicle needs accurate localization method to obtain global positions. As stated earlier in the Introduction, current localization methods provide vehicle positioning information with centimeter-level accuracy. The localization results may be GPS-type positions with latitude, longitude and height values based on the earth geodetic coordinate. The GPS-type positions for all vehicles need to be transformed into cartesian coordinates with the same coordinate origin to be used in the inter-vehicular distance approximation algorithm. The approximation algorithm neglects the height information and assumes two-dimensional flat road surface. With actual height changes of the road, the actual inter-vehicular distances are larger than the approximated values. This makes vehicle following safer since the vehicles actually keep larger gaps between them.

In Fig.~\ref{fig:distance}, the triangles represent the localized positions given by the localization results, and the squares represent the vehicles' true positions which are the positions of the GPS receivers on the vehicles. Via inter-vehicular wireless communication, the follower vehicle $i$ receives the localized position of its preceding vehicle ($x_{i-1}$, $y_{i-1}$). Together with the localized position of itself ($x_i$, $y_i$) and the digital map with lane center points, the follower vehicle obtains the approximated inter-vehicular distance using the following procedure.

Firstly, on the cartesian coordinate system, a bounding box is used to select a rectangular region that covers the localized positions of both vehicles. The bounding box is selected such that the smallest distances from the localized positions to the sides of the bounding box (shown by the dashed lines in Fig.~\ref{fig:distance}) are all the same. The bounding box covers a segment of the road with a number of lane center points. The lane center points covered by the bounding box are fitted with a quadratic curve function. The localized positions may not be exactly on the fitted quadratic curve of lane center points due to localization errors or that the vehicles may not follow the lane center exactly. Thus, the localized positions are projected on the quadratic curve. A projection point is obtained as the intersection between the quadratic curve and the straight line that passes through the localized position and is perpendicular to the two lane center points closest to the localized position. Two projection points are obtained for the localized positions of the two vehicles.

The approximated inter-vehicular distance is defined as the arc length between the two projection points on the quadratic curve, see the thicker black curve that overlaps some lane center points in Fig.~\ref{fig:distance}. Assuming the quadratic curve function as

\begin{equation}
\begin{split}
y = ax^2 + bx + c
\end{split}
\end{equation}
where $a$, $b$, and $c$ are the coefficients that can be obtained through fitting the lane center points in the bounding box. The arc length $L$ between the two projection points, ($x^*_{i-1}$, $y^*_{i-1}$) for the preceding vehicle and ($x^*_i$, $y^*_i$) for the follower vehicle, on the quadratic curve can be computed as

\begin{equation}
\begin{split}
L = \int_{x^*_{i-1}}^{x^*_i} \sqrt{1+ y'^2} dx = \int_{x^*_{i-1}}^{x^*_i} \sqrt{4a^2x^2 + 4abx + b^2 + 1} dx
\end{split}
\end{equation}

In the event that the lane center points are perfectly on a straight line, the coefficient $a$ of the quadratic function becomes zero. The quadratic function actually turns into a line function. However, the above methodology to obtain the approximated inter-vehicular distance still works by setting $a = 0$.

\section{Robotic Emulation of Traffic}

We created the robotic emulation of traffic using mobile robots to evaluate the inter-vehicular approximation algorithm with robot following experiments. The robotic emulation of traffic is both computationally and financially affordable as compared to real-world connected and automated vehicles and allows us to proof-test algorithms in a faster manner. In the following, we introduce the mobile robot testbed preparation which include the robot hardware and software, system identification of robot longitudinal dynamics, robot wireless communication, robot lane keeping, in-lab emulated city and emulated GPS, and robot self-localization.

\subsection{Robot Hardware and Software}

\begin{figure}[htbp]
\centering
\includegraphics[width=3.4in]{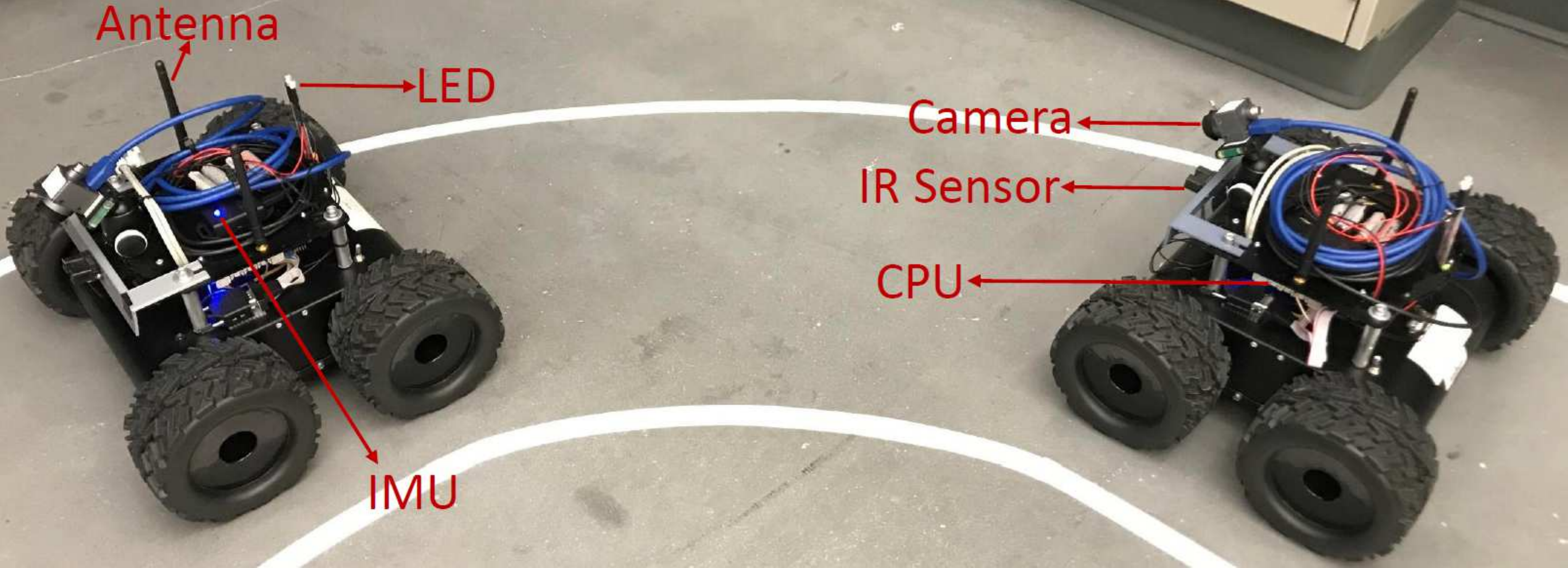}
\caption{Two robots used for CACC robot following experiments.}
\label{fig:robots}
\end{figure}

The mobile robots are differential-drive skid-steering robots, Wifibot Lab V4, developed by Nexter Robotics, see the robots in Fig.~\ref{fig:robots}. Each robot has a mini-computer with Intel Core I5 CPU. The CPU operates a Linux System Ubuntu 14.04. Qt is installed on the Linux System as the C++ IDE. All the robot system capabilities are programmed in C++ for real-time application. The mini-computer connects to a forward-facing camera for lane detection, hosts a wifi card for inter-robot wireless communication, and interfaces with a low-level micro-controller through RS232 serial communication. The mini-computer executes high-level algorithm processing and sends velocity commands to the low-level micro-controller for motor motion control. The low-level micro-controller also obtains wheel encoder reading as robot velocity and Infrared (IR) sensor reading as inter-vehicular distance. Note that we use the IR sensor to emulate radar because the IR sensor is much more affordable and can provide desirable inter-vehicular distance measurements for the robot experiment purpose.

\subsection{Robot Longitudinal Dynamics}

\begin{figure}[htbp]
\centering
\includegraphics[width=3.4in]{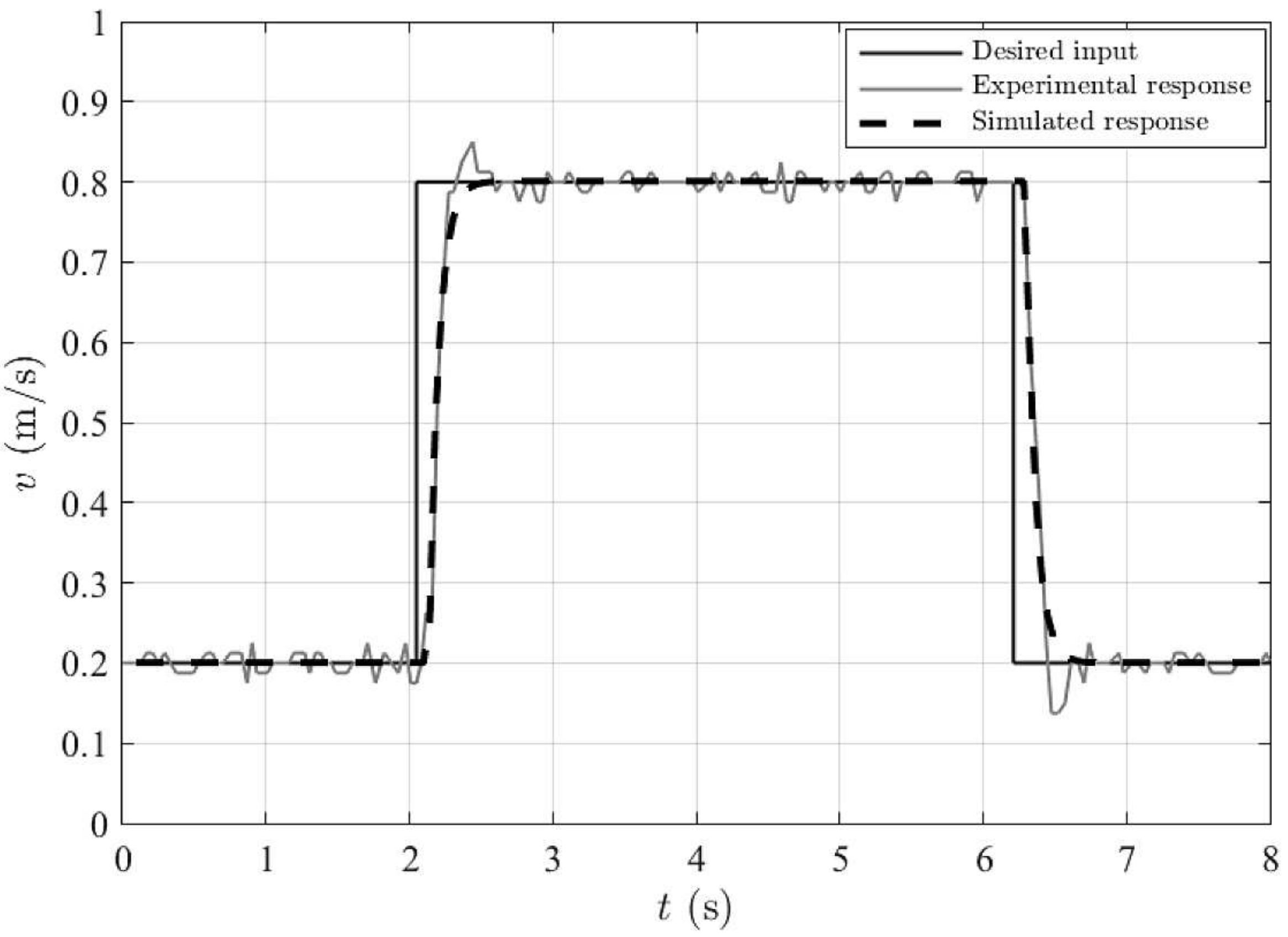}
\caption{System Identification to obtain the transfer function of robot longitudinal dynamics.}
\label{fig:system_id}
\end{figure}

The CACC system requires the transfer function of the mobile robot longitudinal dynamics $G$. Therefore, we send step inputs of desired velocity to the robot and obtain the actual velocity output using the wheel encoder, see Fig.~\ref{fig:system_id}. Note that the robot takes in only velocity commands for motor control. We consider the transfer function from the desired velocity input to actual velocity ouput as a first-order system $\frac {1} {\tau s + 1} e^{\tau_d s}$ with $s$ as the Laplace Transform variable, $\tau$ as the time constant of the first order system, and $\tau_d$ as the time delay. Using MATLAB System Identification toolbox, we obtained $\tau$ = 0.0661 seconds and $\tau_d$ = 0.04 seconds for the experimental response data in Fig.~\ref{fig:system_id}. We then use Simulink to obtain the simulated response of the obtained first-order system to the same desired step input. In Fig.~\ref{fig:system_id}, the simulated response matches the actual velocity output fairly well.

As the robot longitudinal dynamics transfer function $G$ in the CACC system requires the desired input as acceleration and output as traversed distance, we obtain $G$ as
\begin{equation}
\begin{split}
G = \frac {1}{s^2(\tau s + 1)} e^{\tau_d s}
\end{split}
\end{equation}

\subsection{Wireless Communication}

The robots utilize wifi for inter-robot wireless communication. The mini-computer on each robot has a 802.11a/b/g Qualcomm Atheros AR93xx WLAN interface card and an antenna, see Fig.~\ref{fig:robots}. The wireless communication is realized through User Datagram Protocol (UDP) programmed in C++. The communication is decentralized ad hoc network with direct robot-to-robot communication. The communication spectrum is in the 5.9GHz frequency band with sending and receiving rate fixed at 20Hz. In-lab communication tests showed very little (less than 5\%) packet loss.

In the CACC robot experiments, each robot has a unique IP address and pre-stores its predecessor and follower's IP addresses. In this manner, a preceding robot sends its information exclusively to its follower robot using the follower robot's IP address. We acknowledge that this approach may not be feasible in real-world driving. However, it's possible to have a relative positioning algorithm to identify surrounding vehicles as long as they are engaged in the current vehicle following activity. This needs to be further investigated and implemented in future work.

\subsection{Lane Keeping}

Lane keeping capability was developed for the mobile robots to run on in-lab artificial lanes autonomously. As the lane keeping method is documented in our previous work in details \cite{lin2018integrating}, we provide a description of the method here. The lane keeping includes lane detection using computer vision and lane following using pure pursuit path following control. For lane detection, a forward-facing camera is used to obtain images of lane markers in front of the robot. The computer vision algorithm includes undistorting each image through camera calibration, extracting lane markers via edge detection, obtaining a top view of the lane markers through inverse perspective mapping, and removing outliers (erroneous lane markers) by checking lane width and lane slope. The middle point of the detected left and right lane marker points for each row within a defined region of interest is obtained to form the center of lane. The computer vision algorithm is programmed in C++ with OpenCV libraries.

Pure pursuit control is used for a mobile robot to follow the obtained center of lane. The pure pursuit control algorithm utilizes a constant look-ahead horizontal line to intersect the center of lane to obtain a look-ahead point. An instantaneous turning circle that connects the look-ahead point and the vehicle centroid, and whose circle center is on the line that is perpendicular to the robot body length direction can then be constructed. The radius of the instantaneous turning circle is the desired instantaneous turning radius for the robot. Given a target longitudinal velocity provided by the CACC control system, the left and right wheel velocities of the robot can be computed using robot kinematics to achieve the desired instantaneous turning radius. For the mathematical details, readers can refer to our previous work \cite{lin2018integrating}. Note that the pure pursuit control does not guarantee that the robot stays exactly on the center of lane since there is no lane deviation feedback control.

%(sigle image based, from pixels to ground truth)

% scale down version comparing robot with real cars

%\IEEEpubidadjcol

\subsection{In-lab Track \& Emulated GPS}

We built an in-lab emulated city with artificial lanes to emulate road infrastructure. Fig.~\ref{fig:localization} in the CACC Experiments and Results section shows a top view of the in-lab emulated city from an overhead camera. The emulated city has an intersection in the middle and surrounding lane tracks on the outside. The robot following experiments were conducted on the very outside closed-loop track with solid lane markers. On the top-view image, we manually obtained lane center points for the outside track to create the digital map needed for the inter-vehicular distance approximation algorithm. The average distance between two lane center points is around 15 centimeters on the robot track.

A vision-based emulated GPS is developed using the overhead camera to provide robot positions in the emulated city. The main idea of the emulated GPS is to automatically identify and localize a LED light placed on top of each robot, see Fig.~\ref{fig:robots}. The corresponding computer vision algorithm includes undistorting the raw image obtained from the overhead camera, setting a brightness threshold on the raw image to obtain a black-and-white image whose white pixels represent the bright LED light, and obtaining the robot horizontal and vertical positions as the average of all the white pixels' horizontal and vertical locations on the image, respectively. The obtained horizontal and vertical positions are cartesian coordinates with the bottom left corner of the undistorted image as the coordinate origin. The positions are originally in the unit of pixels but are converted to values in meters by comparing the same object length in pixels on the image and in meters of ground truth.

The emulated GPS is run on an independent laptop with a Linux system for real-time image processing since the overhead camera image size is large (5Mb). The independent laptop also has wifi which allows the position values to be sent to the corresponding robots instantaneously using the UDP introduced in the Wireless Communication session. The emulated GPS provides positions at 2Hz since the overhead camera frame rate is 2 frames per second. The emulated GPS also outputs heading angle value which is computed as the angle from a horizontal line to the line that connects two neighboring positions of the same robot obtained from two neighboring image frames.

%(Need to discuss how GPS differentiates different robots, positions provided by the GPS are considered highly accurate due to the method they are obtained.)

\subsection{Localization}

The localization provides localized positions of the robots to be used in the inter-vehicular distance approximation algorithm. The emulated GPS provides positions only at 2Hz, which is not sufficient for continuous operation of the approximation algorithm. Thus, we fuse the emulated GPS and IMU data using an extended Kalman filter to provide positioning information at the IMU frequency which is $f$ = 100Hz. We achknowledge that the state-of-the-art localization methods for self-driving cars fuse information from more sources such as lidar, camera, odometry, and/or a HD map to provide highly accurate positioning information. However, in our indoor lab setting, fusing just the emulated GPS and IMU data provides desirable accuracy for the approximation algorithm.

\begin{figure}[htbp]
\centering
\includegraphics[width=3in]{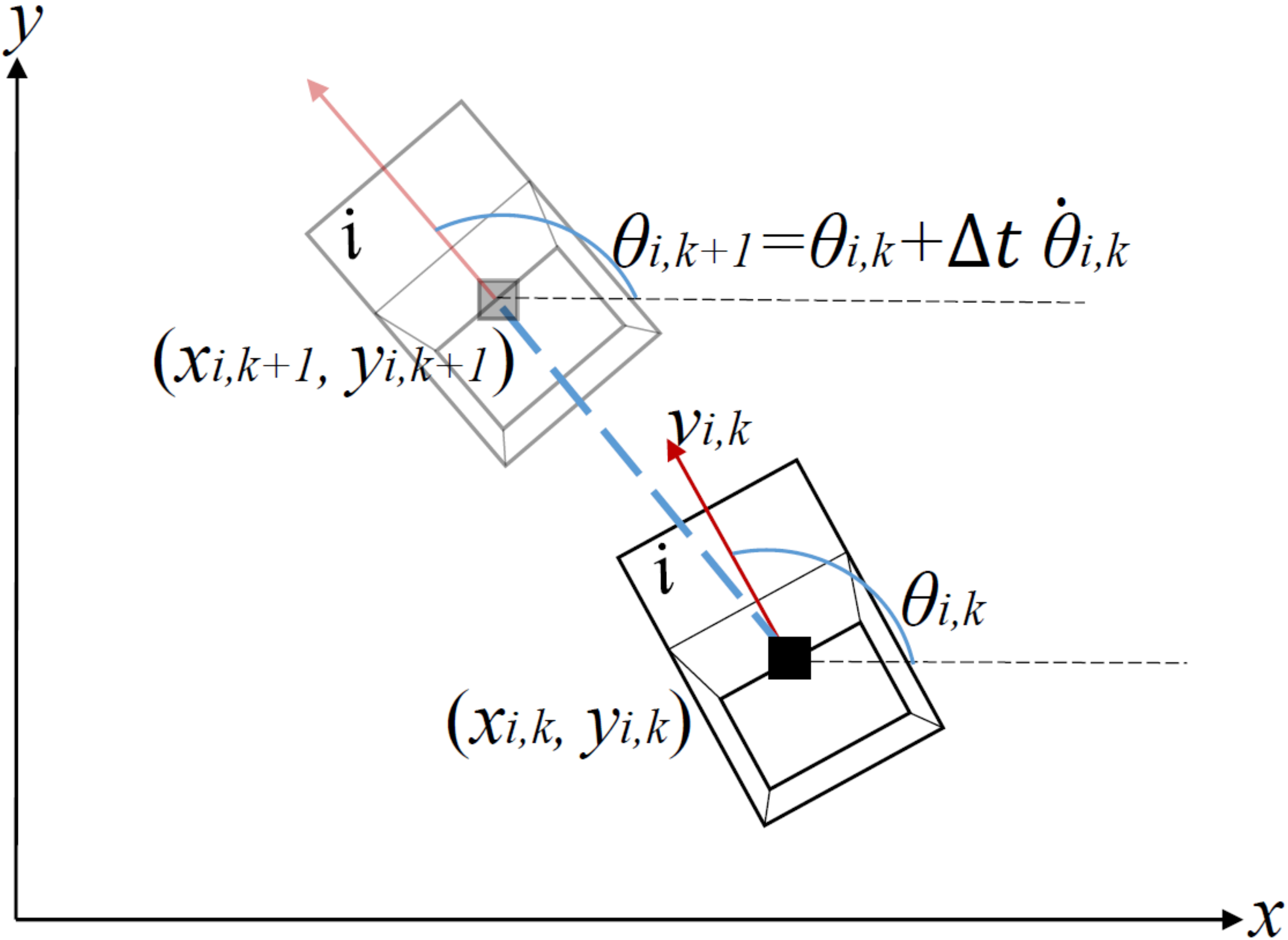}
\caption{Robot motion update from time step $k$ to $k+1$.}
\label{fig:motion}
\end{figure}

In the following, the extended Kalman filter method to fuse the emulated GPS and IMU data is explained. The state equations for the vehicle motion update are
\begin{equation}
\begin{split}
x_{i,k+1} = x_{i,k} + \Delta t v_{i,k} \cos(\theta_{i,k} + \Delta t \dot{\theta}_{i,k}) \\
y_{i,k+1} = y_{i,k} + \Delta t v_{i,k} \sin(\theta_{i,k} + \Delta t \dot{\theta}_{i,k}) \\
\theta_{i,k+1} = \theta_{i,k} + \Delta t \dot{\theta}_{i,k}
\end{split}
\end{equation}
where $x_{i,k+1}$, $y_{i,k+1}$, and $\theta_{i,k+1}$ are the horizontal position, vertical position, and heading angle of the robot, respectively, see Fig.~\ref{fig:motion}. These three variables are also called state variables. The $v_{i,k}$ is the longitudinal velocity magnitude of the robot obtained through wheel encoders. The $\dot{\theta}_{i,k}$ is the yaw rate (angular velocity) obtained through the IMU. The $\Delta t$ is the updating time step which is determined by the IMU frequency $\Delta t = 1/f$ = 0.01 seconds.

The measurement variables for the correction step in Kalman filtering are the same as the state variables and are provided by the emulated GPS. Using state space representation, we rewrite the state and measurement update equations in vector form as
\begin{equation}
\begin{split}
X_{i,k+1} = f(X_{i,k},U_{i,k}) + W \\
Z_{i,k+1} = H X_{i,k+1} + V \\
\end{split}
\end{equation}
where $X_{i,k+1} = [x_{i,k+1}, y_{i,k+1}, \theta_{i,k+1}]^\intercal$ is the vector of the state variables; $U_{i,k} = [v_{i,k}, \dot{\theta}_{i,k}]^\intercal$ is the vector of the inputs which include the longitudinal velocity $v_{i,k}$ obtained from wheel encoders and the yaw rate $\dot{\theta}_{i,k}$ obtained from the IMU; $Z_{i,k+1} = [x^{GPS}_{i,k+1}, y^{GPS}_{i,k+1}, \theta^{GPS}_{i,k+1}]^\intercal$ is the vector of the measurement variables. The $H$ matrix is
\[
H=
  \begin{bmatrix}
    1 & 0 & 0 \\
    0 & 1 & 0 \\
    0 & 0 & 1
  \end{bmatrix}
\]

The $W$ is the process noise covariance matrix that results from the inaccuracy of the state equations. In the state equations, the vehicle longitudinal velocity instead of the actual velocity is used which results in the inaccuracy. In other words, the actual velocity should have been used in the state equations to describe the correct vehicle motion update. The actual velocity consists of both longitudinal and lateral velocity components where the lateral velocity contributes to the vehicle turning behavior. As wheel encoder sensors measure just the longitudinal velocity, we use the longitudinal velocity in the state equations. Due to the inaccuracy, we define the constant $W$ as
\[
W=
  \begin{bmatrix}
    0.01 & 0 & 0 \\
    0 & 0.01 & 0 \\
    0 & 0 & 0.001
  \end{bmatrix}
\]

The $V$ is the measurement noise covariance matrix that results from the inaccuracy of the measurements. As our vision-based algorithm delivers highly accurate emulated GPS positions, we consider zero error for the position measurements. Thus, we define the constant $V$ as
\[
V=
  \begin{bmatrix}
    0 & 0 & 0 \\
    0 & 0 & 0 \\
    0 & 0 & 0.01
  \end{bmatrix}
\]

To use the extended Kalman filter, the state transition matrix $A$ is obtained as the Jacobian of the non-linear state equations.
\[
A=
  \begin{bmatrix}
    1 & 0 & -v_{i,k} \Delta t \sin(\theta_{i,k} + \Delta t \dot{\theta}_{i,k}) \\
    0 & 1 & v_{i,k} \Delta t \cos(\theta_{i,k} + \Delta t \dot{\theta}_{i,k}) \\
    0 & 0 & 1
  \end{bmatrix}
\]

The following shows the extended Kalman filter update steps. First, a prior error covariance estimate of the next time step $k+1$ is computed as
\begin{equation}
\begin{split}
P^-_{k+1} = A P_k A^\intercal + Q
\end{split}
\end{equation}
where $P_k$ is the posterior error covariance estimate for the current time step $k$; its initial value is defined as
\[
P_0=
  \begin{bmatrix}
    0.01 & 0 & 0 \\
    0 & 0.01 & 0 \\
    0 & 0 & 0.01
  \end{bmatrix}
\]

The Kalman filter gain is calculated as
\begin{equation}
\begin{split}
K_{k+1} = \frac{P^-_{k+1} H^\intercal}{H P^-_{k+1} H^\intercal + R}
\end{split}
\end{equation}

The posterior state estimate can then be computed as
\begin{equation}
\begin{split}
X_{i,k+1} = f(X_{i,k},U_{i,k}) + K_{k+1} [Z_{i,k} - H f(X_{i,k},U_{i,k})]
\end{split}
\end{equation}
These are the state variable outputs among which the robot horizontal and vertical positions are used in the inter-vehicular distance approximation algorithm.

Last, the posterior error covariance estimate $P_{k+1}$ is updated as
\begin{equation}
\begin{split}
P_{k+1} = (I - K_{k+1} H) P^-_{k+1}
\end{split}
\end{equation}
where I is a 3 by 3 identity matrix.

Note that the above extended Kalman filter update steps are only used when new emulated GPS measurements are available. When new emulated GPS measurements are not available, the state variable outputs are computed using just the robot motion update state equations $X_{i,k+1} = f(X_{i,k},U_{i,k})$. Thus, the robot positions for the time periods between the arrivals of two emulated GPS positions rely on the IMU data. The matrix calculation for the Kalman filtering was realized with Armadillo: a template-based C++ library for linear algebra \cite{sanderson2016armadillo,sanderson2018user}.

%\subsubsection{}
%Subsubsection text here.

\section{CACC Experiments and Results}

With the mobile robot testbed, we conducted two sets of robot following experiments to evaluate the proposed inter-vehicular distance approximation algorithm. For the first set of experiment, the inter-robot distance was provided solely by the proposed approximation algorithm. In other words, no range sensor (IR sensor) was used in the first set of experiment. For the second set of experiment, the range sensor (IR sensor) was used together with the proposed approximation algorithm to provide the inter-robot distance. In the second set of experiment, the IR sensor provided the inter-robot distance whenever possible and the proposed approximation algorithm was used only when the IR sensor was not able to provide the inter-robot distance around the curve. The second set of experiment is based on a switching mechanism that allows a vehicle to switch from using the range sensor to using the proposed approximation algorithm to obtain the inter-vehicular distance. The switching mechanism design can be potentially applied on the current commercial ACC systems so that it allows a vehicle to continue to follow during target detection loss. In the following, the two sets of robot experiments are explained and the corresponding results are shown.

\subsection{CACC without range sensor}

\begin{figure}
    \centering
    \subfigure[]{
    \includegraphics[width=3.3in]{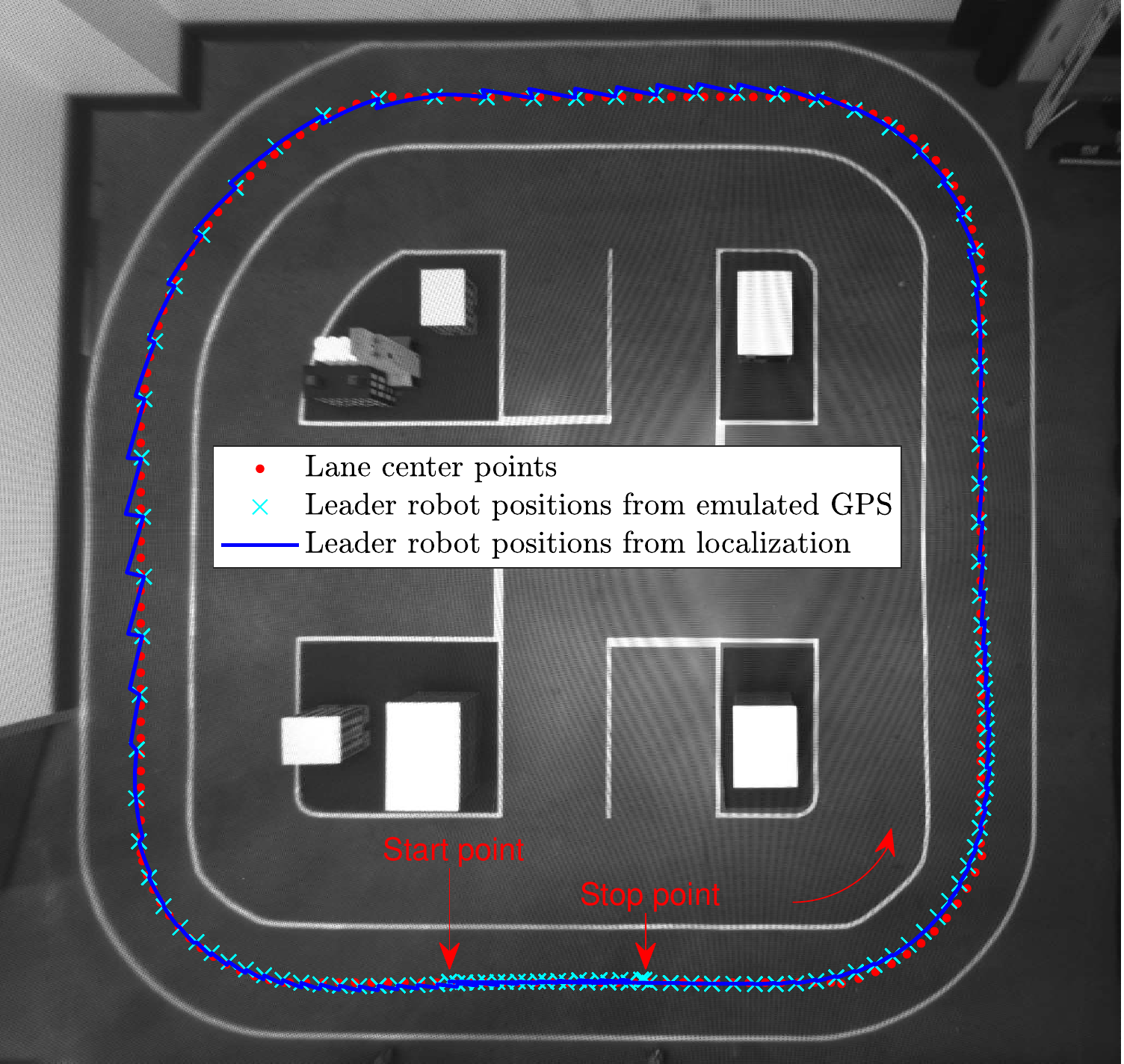}
    \label{fig:loc106}
    }
    \subfigure[]{
    \includegraphics[width=3.3in]{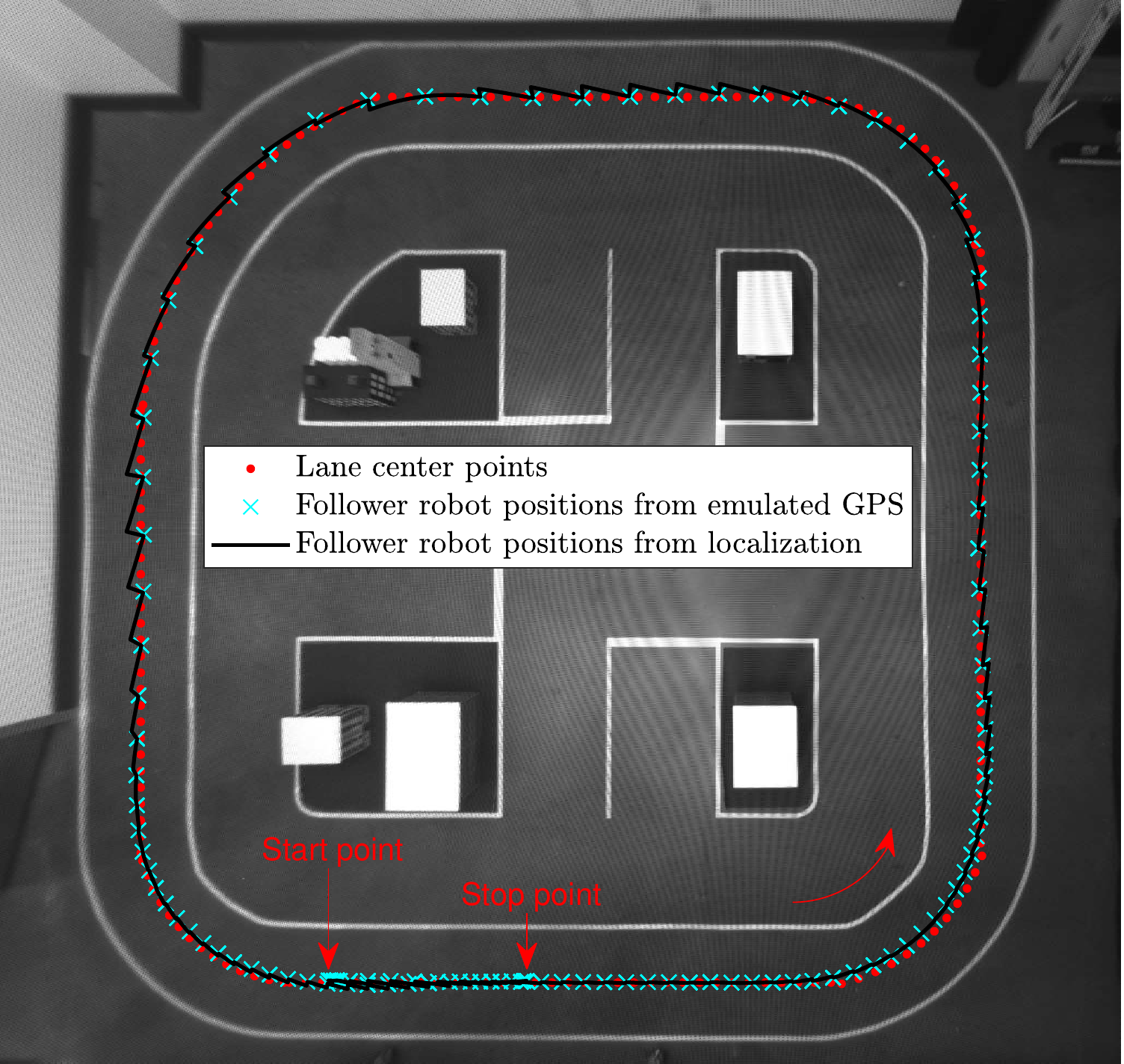}
    \label{fig:loc107}
    }
    \caption{Localization results for CACC using solely the proposed approximation algorithm (without using the IR sensor). \subref{fig:loc106} leader and \subref{fig:loc107} follower robot localization results.}\label{fig:localization}
\end{figure}

\begin{figure}
    \centering
    \subfigure[]{
    \includegraphics[width=3.3in]{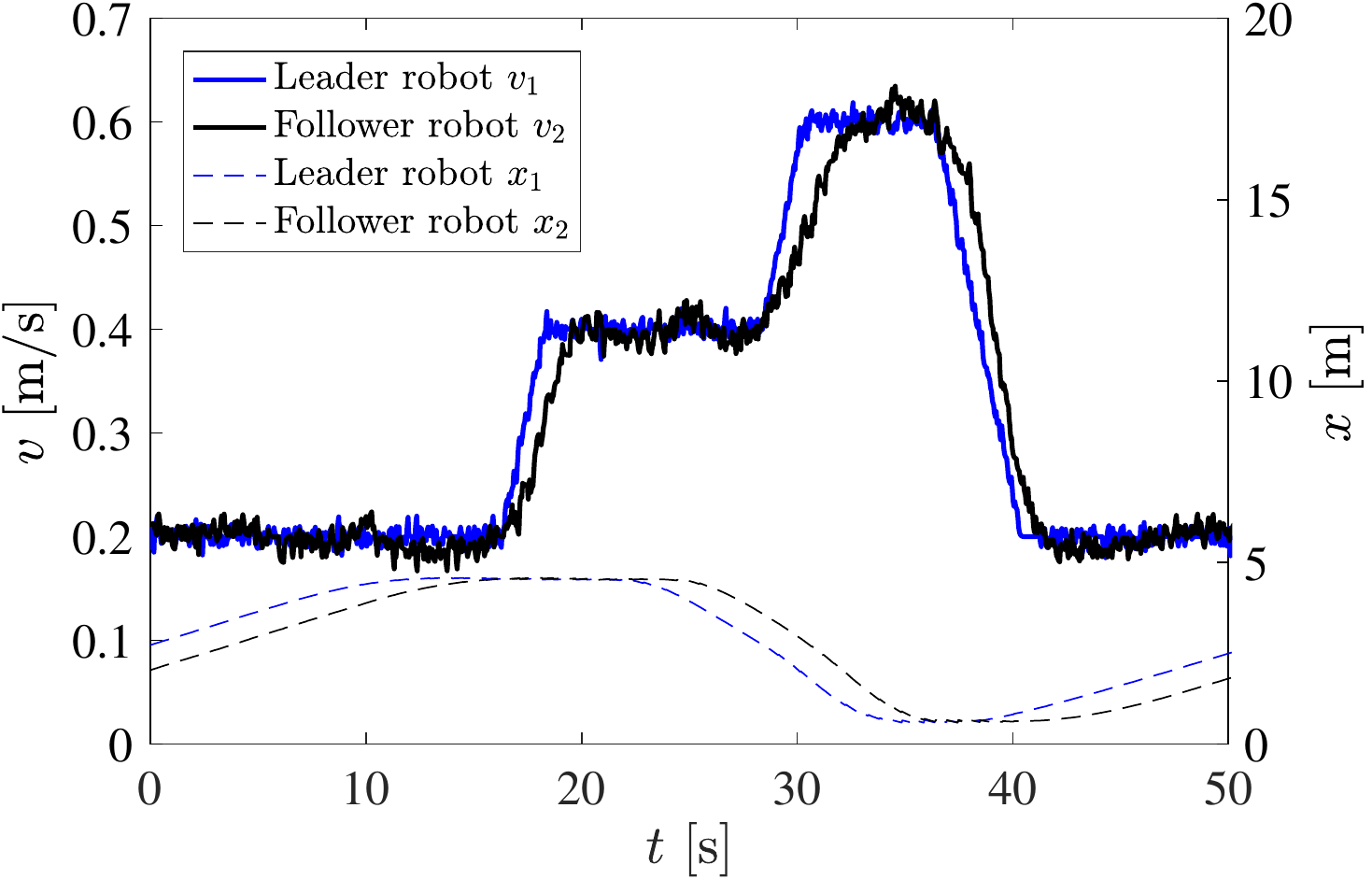}
    \label{fig:v}
    }
    \subfigure[]{
    \includegraphics[width=3.3in]{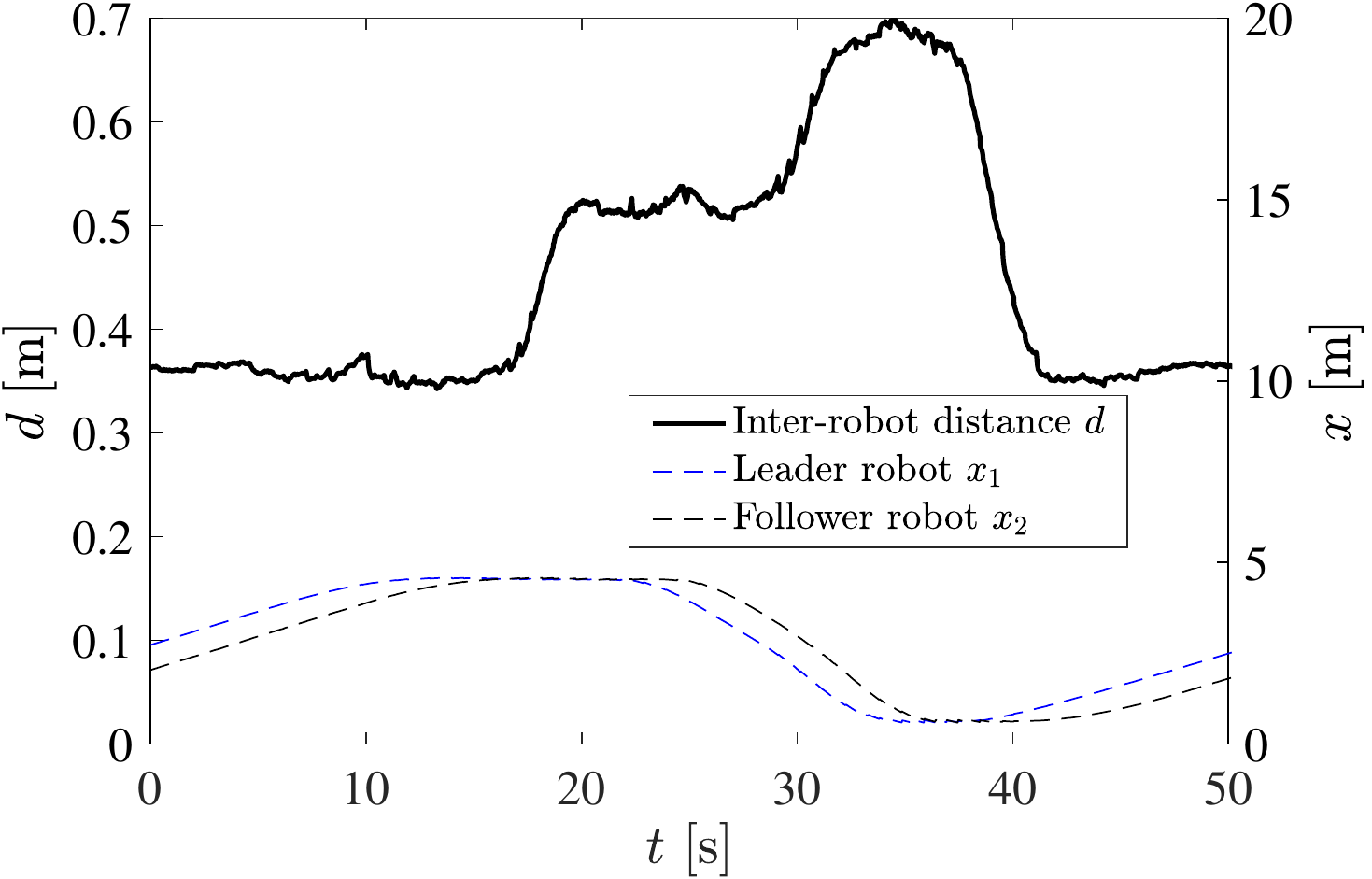}
    \label{fig:d}
    }
    \subfigure[]{
    \includegraphics[width=3.3in]{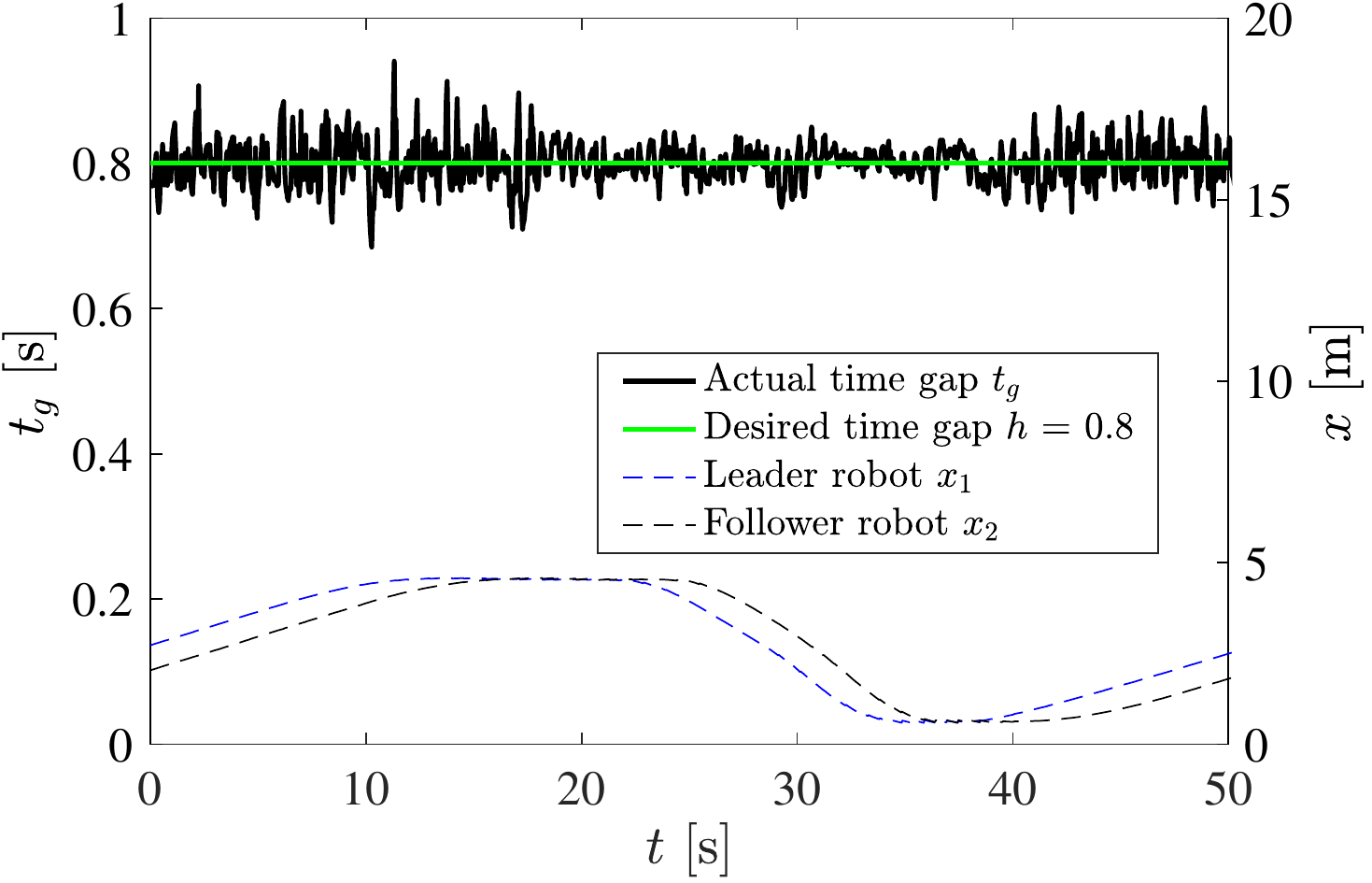}
    \label{fig:tg}
    }
    \caption{Robot following experiment results for CACC using solely the proposed approximation algorithm (without using the IR sensor). \subref{fig:v} velocity, \subref{fig:d} inter-vehicular distance, and \subref{fig:tg} time gap. The two dashed curves at the bottom in each plot show the horizontal positions of the two robots.}\label{fig:cacc_results}
\end{figure}

In the first set of robot following experiment, a follower robot obtained the inter-robot distance using solely the proposed inter-vehicular distance approximation algorithm during the entire experiment and the IR sensor was not used at all. Two robots were used with one being the leader robot and the other being the follower robot. The leader robot's velocity profile was pre-defined and pre-programmed such that it traversed a complete circle of the outside track of the emulated city. We considered various speeds in one run to investigate the impact of speeds on the accuracy of the inter-vehicular distance approximation, see Fig.~\ref{fig:v}. A video of the robot following experiment can be seen on the website of the Autonomous Systems and Intelligent Machines Lab in the Mechanical Engineering Department at Virginia Tech.

Fig.~\ref{fig:localization} shows the localization results for both robots using the robot localization method introduced previously. Comparing the emulated GPS positions and the lane center points of the digital map, we conclude that the lane keeping method desirably restricted the robot to follow the center of lane. However, The robots were not exactly on the center of lane, especially around the curve where the robots seemed to slightly ``over-steer'' and stayed inside the circle of the center of lane. This could lead to a slight increase of the follower robot's velocity when the leader robot was at the curve and a slight decrease of the follower robot's velocity when the follower robot was at the curve, since the ``over-steering'' created shorter driving path compared to the lane center path. These can be validated in the velocity plots of Fig.~\ref{fig:v} and Fig.~\ref{fig:v2}.

In Fig.~\ref{fig:localization}, the distance between two emulated GPS positions increases as the robot speed increases. This should also be true to the localized positions obtained from localization. The localization method provides robot positions at a much higher frequency 100Hz compared to the emulated GPS frequency 2Hz. In the plots, the localization results are plotted using curves instead of discreet points due to visualization concerns. Since the latest localized positions are used to estimate the distance in the algorithm, a higher frequency for localization update is essential for smooth distance approximation. For the robot speeds considered, the present localization update frequency 100Hz is sufficient to obtain smooth inter-vehicular distance, see Fig.~\ref{fig:d}.

In Fig.~\ref{fig:tg}, the actual time gap values are very close to the desired time gap $h$ = 0.8 seconds despite the fluctuations mainly caused by the data measurement errors. The actual time gap values have an average of 0.8000 seconds and a standard deviation of 0.0264 seconds. This indicates that the CACC system leads to satisfying robot gap-keeping control.

\subsection{CACC with switching mechanism}

\begin{figure}[htbp]
\centering
\includegraphics[width=3.4in]{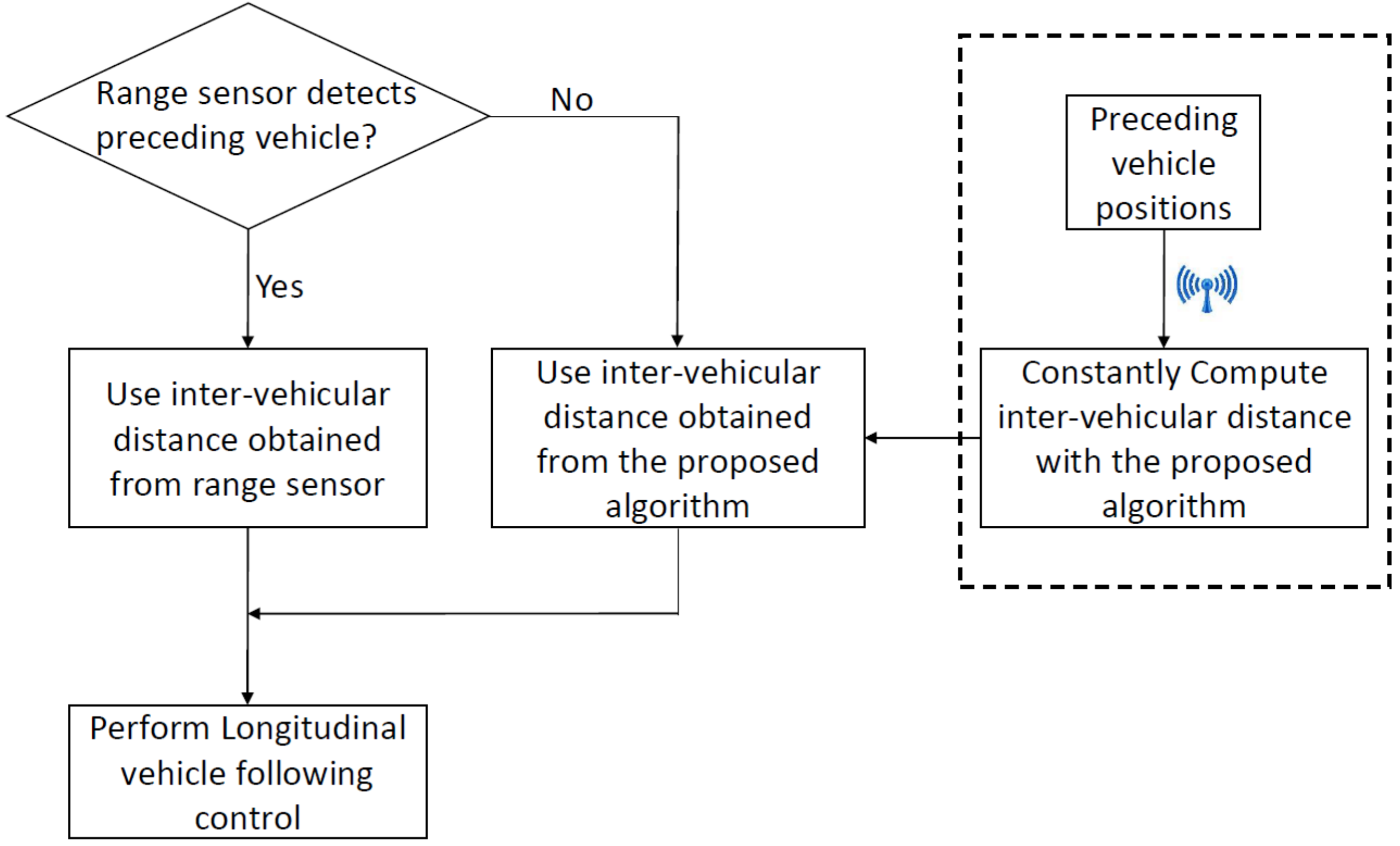}
\caption{Flow chart for the switching mechanism to obtain inter-vehicular distance for CACC.}
\label{fig:flow_chart}
\end{figure}

In the second set of robot following experiment, the follower robot used the range sensor (IR sensor) as the priority option and switched to using the proposed approximation algorithm to obtain the inter-robot distance only when the IR sensor was not able to provide the distance values. Fig.~\ref{fig:flow_chart} shows the flow chart for the switching mechanism. Note that the proposed approximation algorithm computes the inter-vehicular distance all the time in the background, while the computed distances are only used then range sensor is not able to provide the inter-vehicular distance.

\begin{figure}
    \centering
    \subfigure[]{
    \includegraphics[width=3.3in]{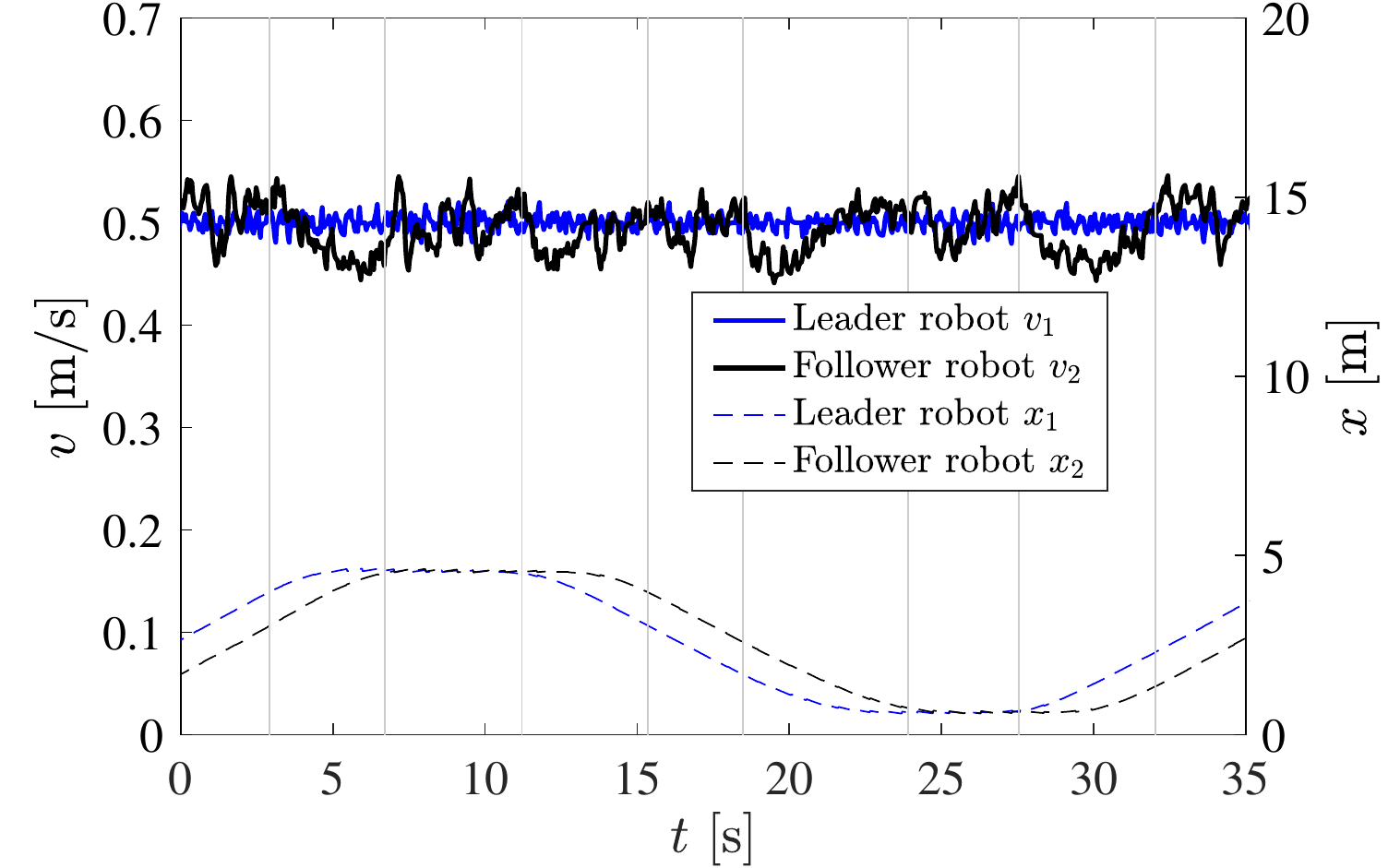}
    \label{fig:v2}
    }
    \subfigure[]{
    \includegraphics[width=3.3in]{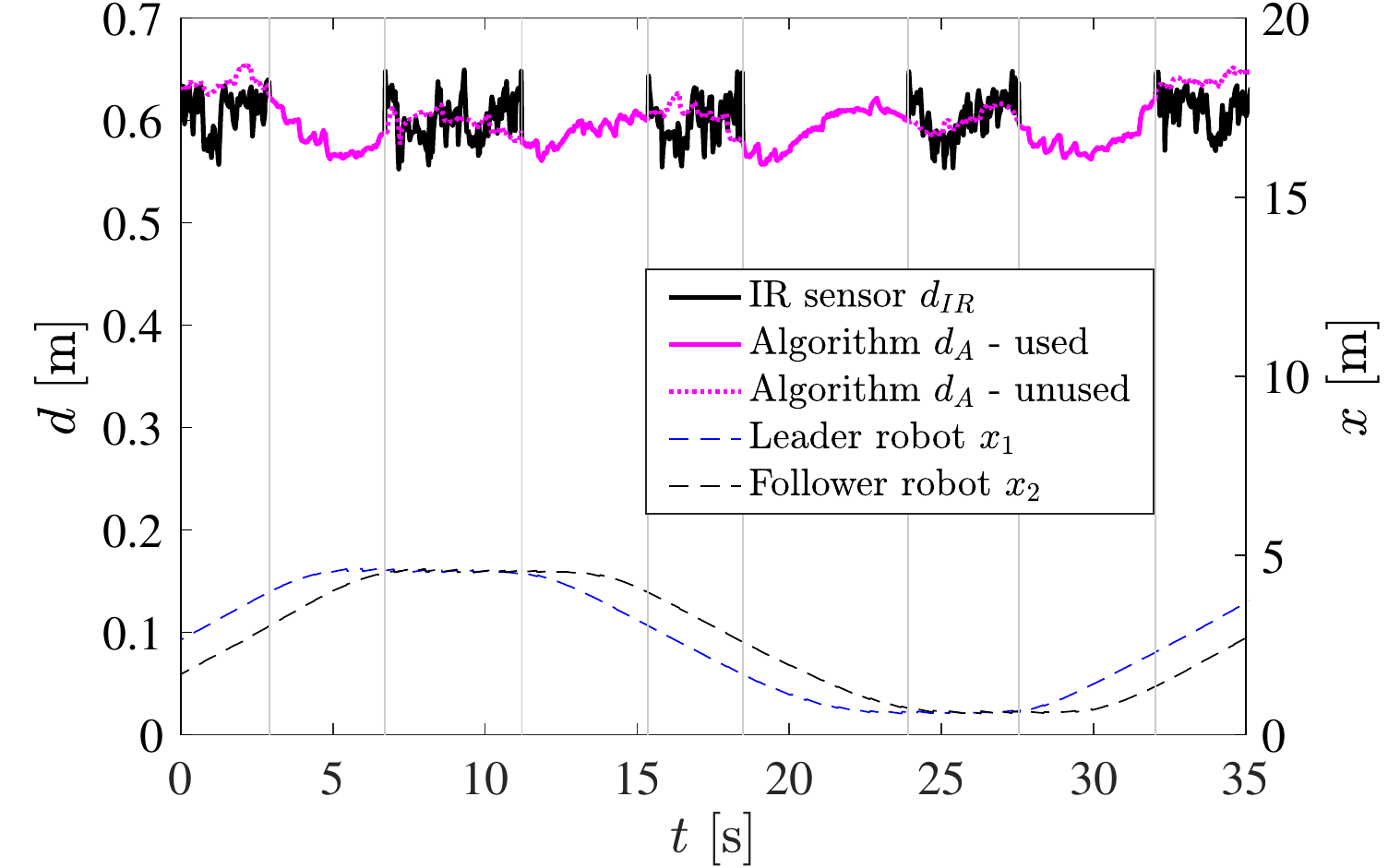}
    \label{fig:d2}
    }
    \subfigure[]{
    \includegraphics[width=3.3in]{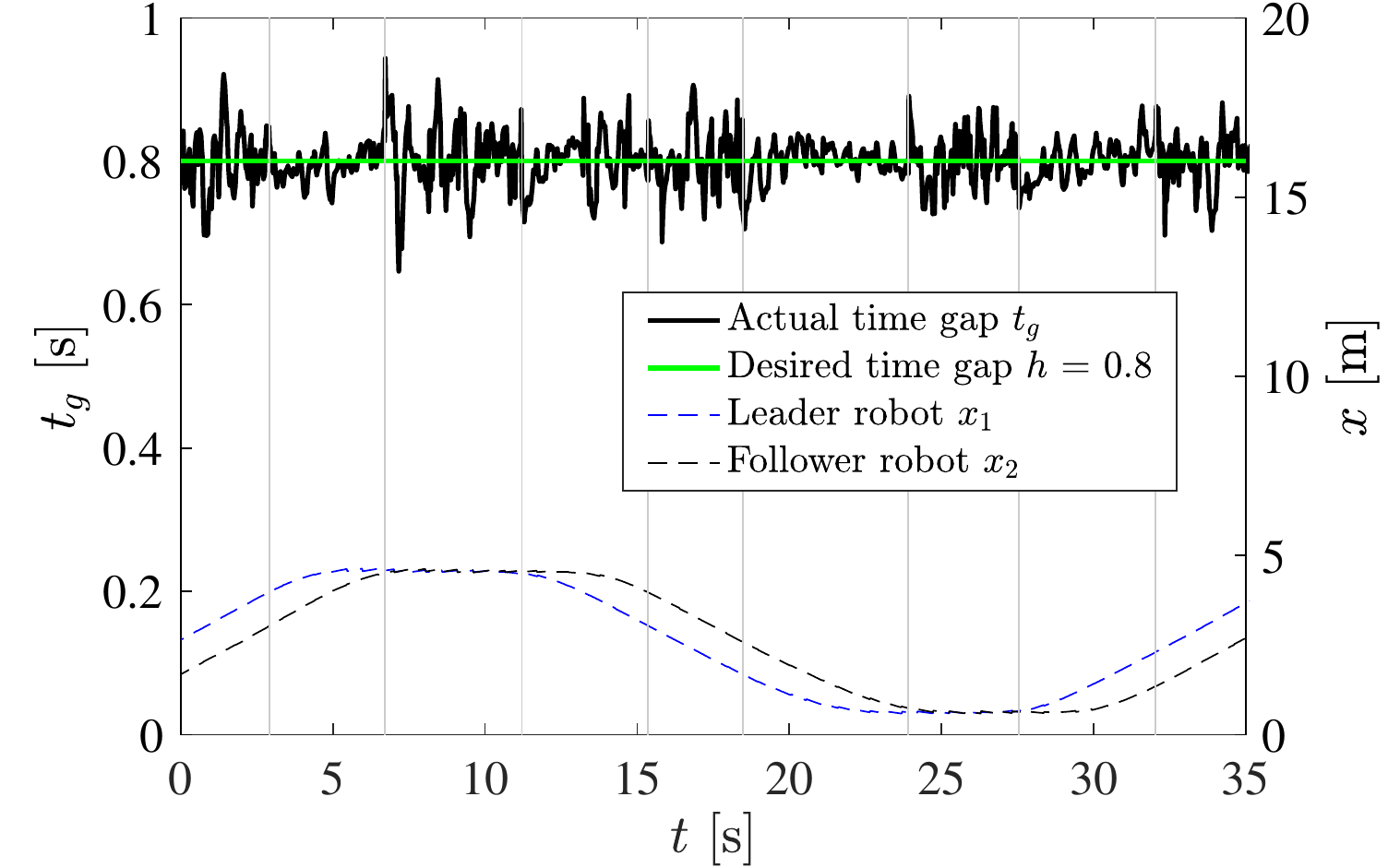}
    \label{fig:tg2}
    }
    \caption{Robot following experiment results for CACC with switching between using the IR sensor and using the proposed algorithm to obtain the inter-robot distance. \subref{fig:v2} velocity, \subref{fig:d2} inter-robot distance, and \subref{fig:tg2} time gap. The two dashed curves at the bottom in each plot show the horizontal positions of the two robots. The grey color vertical lines denotes the times when the switching happened.}\label{fig:cacc_results2}
\end{figure}

For the current automobile ACC systems, the sensors may lose detection of the preceding vehicle due to various reasons at various occasions. For our robot experiment, the occasion of driving around the curve is considered. Since the IR sensor on the robot provides just a scalar representing the distance, we define the switching condition as that the IR sensor returns erroneous distance values. Since the leader robot maintained a constant speed $v$ = 0.5m/s in this experiment (Fig.~\ref{fig:v2}), the IR sensor distance became erroneous when it was significantly larger or smaller than the constant desired distance $h \dot{l}_i + l_0$ = 0.8*0.5 + 0.2 = 0.6m. In fact, we defined the erroneous values as $d < 0.55$ and $d > 0.65$ meters. In other words, the follower robot used the IR sensor to obtain the inter-robot distance when $0.55 \geq d \leq 0.65$ and used the proposed approximation algorithm otherwise. Note that the switching conditions for real cars are different since the existing automobile ACC systems self-determine target detection loss at various occasions. The goal of our robot experiment is to just introduce the switching method.

Fig.~\ref{fig:v2} shows the velocity results of both robots during the experiment. The follower robot experienced velocity fluctuations due to the ``over-steering'' around the curve explained previously. Fig.~\ref{fig:d2} shows the inter-robot distance. By observing the robot horizontal positions, we see that the follower robot utilized the IR sensor to obtain the inter-robot distance when both robots were on straight paths and utilized the proposed approximation algorithm otherwise (around the curve). When the IR sensor distance was used, the proposed approximation algorithm also computed the approximated distance in the background although it was not used. By comparing the IR sensor distance and the unused approximated distance, we conclude that the proposed approximation algorithm provided close approximation to the IR sensor distance. Fig.~\ref{fig:tg2} shows that the actual time gap values of the robot following. The actual time gap values are also close to the desired time gap $h$ = 0.8 seconds with an average of 0.8005 seconds and a standard deviation of 0.0359 seconds.

\section{Conclusion}

In conclusion, we proposed an inter-vehicular distance approximation algorithm for vehicle following with target detection loss. The algorithm integrates inter-vehicular wireless communication, accurate vehicle localization, and a digital map with lane center information. More specifically, a follower vehicle receives the position of its preceding vehicle via wireless communication and utilizes its own position and road geometry from the map to mathematically compute the inter-vehicular distance. We have demonstrated through robot experiments that the proposed algorithm could be a complete replacement of range sensors for long-period vehicle following. We also designed the switching mechanism such that an ACC or CACC system can switch from using a range sensor to using the proposed algorithm to obtain the inter-vehicular distance for short-term target detection loss.

Through the robot following experiments, we discovered some factors that contribute to accurate and smooth inter-vehicular distance approximation using the proposed algorithm. A high vehicle localization update rate and a high wireless communication frequency are essential in the approximation since a follower vehicle requires instantaneous positions of both itself and its preceding vehicle to obtain the instantaneous inter-vehicular distance. The accuracy of the vehicle positions produced by the vehicle localization methods and the correctness of lane center locations of the digital map also directly impacts the approximation accuracy.

In the event of low localization update rates and communication frequencies or communication packet loss, an interpolation or prediction method may be needed to generate vehicle positions with high update rates. Future works also include implementing the proposed inter-vehicular approximation algorithm on real cars and conduct high-speed testing.

\section*{Acknowledgment}

The authors would like to thank Chaoxian Wu for providing the C++ code for the UDP wireless communication and Christopher Kappes for helping with the system identification to obtain the robot longitudinal dynamics.

\ifCLASSOPTIONcaptionsoff
  \newpage
\fi



%\bibliographystyle{IEEEtran}

\bibliographystyle{IEEEtran}
\bibliography{references}

%\begin{thebibliography}{1}


%\end{thebibliography}


\begin{IEEEbiography}[{\includegraphics[width=1in,height=1.25in,clip,keepaspectratio]{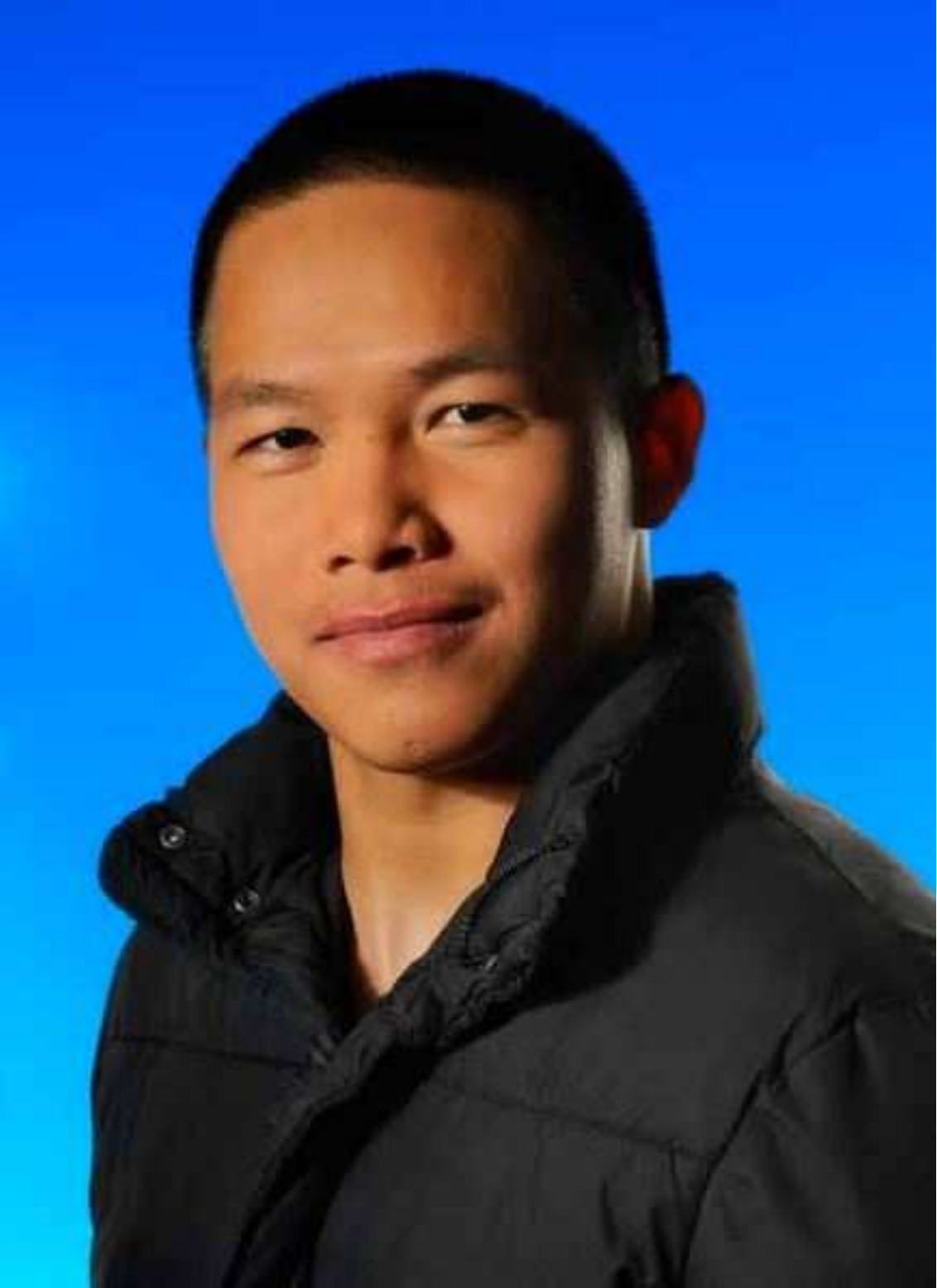}}]{Yuan Lin} received the B.E. degree in Civil Engineering from Nanchang University, China, in 2011 and the Ph.D. degree in Engineering Mechanics from Virginia Tech, Blacksburg, VA, USA, in 2016. After obtaining his PhD, Yuan started to work as a Postdoctoral Research Associate of the Autonomous Systems and Intelligent Machines Lab in the Mechanical Engineering Department at Virginia Tech. His research interests include robotics, control, sensor fusion, and V2X. \end{IEEEbiography}

%\begin{IEEEbiographynophoto}{Azim Eskandarian}
%Dr. Azim Eskandarian is a professor and the Department Head of Mechanical Engineering at Virginia Tech.
%% insert where needed to balance the two columns on the last page with
%%\newpage

\begin{IEEEbiography}[{\includegraphics[width=1in,height=1.25in,clip,keepaspectratio]{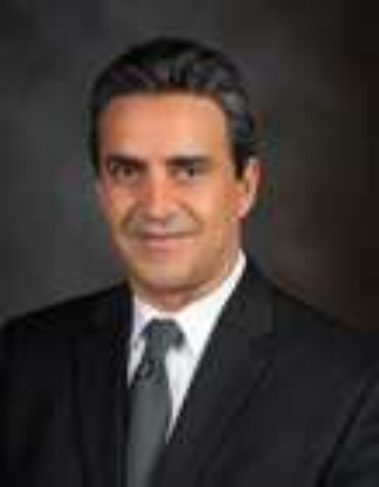}}]{Azim Eskandarian} has been a Professor and Head of the Mechanical Engineering Department at Virginia Tech (VT) since August 2015. He became the Nicholas and Rebecca Des Champs chaired Professor in April 2018. He established the Autonomous Systems and Intelligent Machines laboratory at VT to conduct research in intelligent and autonomous vehicles, and mobile robotics. Prior to that, he was a Professor of Engineering and Applied Science at The George Washington University (GWU) and the founding Director of the Center for Intelligent Systems Research (1996-2015), the director of the “Transportation Safety and Security” University Area of Excellence (2002-2015), and the co-founder of the National Crash Analysis Center (1992) and its Director (1998-2002 \& 5/2013-7/2015). Earlier, he was an Assistant Professor at Pennsylvania State University, York, PA (1989-92) and worked as an engineer/project manager in industry (1983-89). He was awarded the IEEE ITS Society’s Outstanding Researcher Award in 2017 and the GWU’s School of Engineering Outstanding Researcher Award in 2013. Dr. Eskandarian is a fellow of ASME, senior member of IEEE, and member of SAE professional societies. He received his BS, MS, and DSC degrees in Mechanical engineering from GWU, Virginia Tech, and GWU, respectively.\end{IEEEbiography}


%\vfill


\end{document}